\documentclass{article}

\PassOptionsToPackage{round,compress}{natbib}

     \usepackage[preprint]{neurips_2021}

\usepackage[utf8]{inputenc} 
\usepackage[T1]{fontenc}    
\usepackage{hyperref}       
\usepackage{url}            
\usepackage{booktabs}       
\usepackage{amsfonts}       
\usepackage{nicefrac}       
\usepackage{microtype}      
\usepackage{xcolor}         
\usepackage[ruled]{algorithm2e}
\usepackage{amsmath}
\usepackage{amssymb}
\usepackage{multirow}
\usepackage{bbm}
\usepackage{amsthm}
\usepackage{graphicx}
\usepackage{float}
\usepackage{subfigure}

\newtheorem{thm}{Theorem}
\newtheorem{assumption}{Assumption}

\newtheorem{lem}{Lemma}

\title{Federated Learning with Classifier Shift for Class Imbalance}

\author{%
    Yunheng Shen\thanks{Use footnote for providing further information
        about author (webpage, alternative address)---\emph{not} for acknowledging
        funding agencies.} \\
    Department of Automation\\
    Tsinghua University\\
    \texttt{shenyh19@mails.tsinghua.edu.cn} \\
    \And
    Haoxiang Wang\\
    Institute for Interdisciplinary Information Sciences\\
    Tsinghua University\\
    \texttt{wanghaox19@mails.tsinghua.edu.cn} \\
    \And
    Hairong Lv\\
    Department of Automation\\
    Tsinghua University\\
    \texttt{lvhairong@tsinghua.edu.cn} \\
}

\begin{document}

    \maketitle

    \begin{abstract}
        Federated learning aims to learn a global model collaboratively while the training data belongs to different clients and is not allowed to be exchanged. However, the statistical heterogeneity challenge on non-IID data, such as class imbalance in classification, will cause client drift and significantly reduce the performance of the global model. This paper proposes a simple and effective approach named FedShift which adds the shift on the classifier output during the local training phase to alleviate the negative impact of class imbalance. We theoretically prove that the classifier shift in FedShift can make the local optimum consistent with the global optimum and ensure the convergence of the algorithm. Moreover, our experiments indicate that FedShift significantly outperforms the other state-of-the-art federated learning approaches on various datasets regarding accuracy and communication efficiency.

    \end{abstract}

    \section{Introduction}
    There are already numerous edge devices such as smartphones and IoT devices that can collect valuable raw data, and ones expect to use these data to complete some intelligent tasks such as image recognition or text generation. However, deep learning, the most effective algorithm for accomplishing these tasks, requires huge data to train the model, making it challenging to learn a good enough model from the data owned by a single edge device. Besides, due to data privacy, data protection regulations \citep{voigt2017eu}, and the massive overhead of data transmission, it is unrealistic to aggregate data from different clients (edge devices) in a server for training. Therefore, federated learning (FL) \citep{kairouz2019advances} has emerged to solve the problem of jointly learning a global model without sharing the private data.

    Although federated learning has shown good performance in many applications \citep{kaissis2020secure,liu2020fedvision}, there are still several important challenges that require researchers to pay attention to, namely privacy, communication cost, and statistical heterogeneity \citep{ji2021emerging}. Statistical heterogeneity means that client data is non-IID (independent and identically distributed). \citet{zhao2018federated} show that the accuracy of the federated learning algorithm has decreased significantly in the case of non-IID data. There are many methods proposed to address the challenge of statistical heterogeneity. FedProx  \citep{MLSYS2020_38af8613} introduces a proximal term to constrain the update of the local model, and SCAFFOLD \citep{karimireddy2020scaffold} corrects the gradient of each local update to reduce the variance. However, these methods do not bring significant improvement because they only implicitly deal with the fundamental dilemma caused by statistical heterogeneity, that is, the optimal objective of local update is inconsistent with the optimal objective of global update.

    In this work, we propose an approach FedShift to explicitly solve the above fundamental dilemma in the statistical heterogeneity challenge. FedShift is a simple and effective approach which adds the shift on the classifier output calculated by the client category distribution and makes the local optimal models satisfy the global optimum. We also prove the convergence results of FedShift in the strongly convex and non-convex cases and compare with FedAvg, which does not have the classifier shift. Numerous experiments are conducted to evaluate the effectiveness of FedShift, which demonstrate that FedShift outperforms the other state-of-the-art federated learning algorithms in test accuracy and communication efficiency on various datasets, including Cifar10, Cinic10 and Tiny-Imagenet.

    \section{Related Works}

    FedAvg \citep{mcmahan2017communication} is the benchmark method in federated learning, which has demonstrated reliability in image classification and language modeling tasks. Each round of FedAvg mainly contains two phases, client update and server aggregation. First, each client that is selected to participate in training downloads the latest global model from the server, and updates the model locally using stochastic gradient descent. Then, the server collects the updated models from each client and aggregates them to obtain a new global model by averaging the model weights.


    Unlike the privacy and communication challenges, the statistical heterogeneity challenge is a unique and popular issue in the federated learning paradigm \citep{kairouz2019advances}.  According to the two stages of federated learning mentioned above, the contributions of these studies can be roughly divided into local update improvements and aggregation improvements. Our work is an improvement in local update phase, so it can be combined with existing aggregation improvements without any conflict.

    As for the aggregation improvements on non-IID data, there are a series of related studies. PFNM  \citep{yurochkin2019bayesian} and FedMA  \citep{wang2020federated} apply the Bayesian non-parametric mechanism to study the permutation invariance of the neural network, and match the neurons of client neural networks to the global neurons. Moreover, methods such as adaptive weights \citep{yeganeh2020inverse}, attention mechanisms \citep{ji2019learning}, and normalization \citep{wang2020tackling} are also used to improve the aggregation effect for statistical heterogeneity.

    There are also many studies to alleviate the negative effects of statistical heterogeneity in the local update phase. \citet{MLSYS2020_38af8613} propose the FedProx algorithm, which adds a regular term to the loss function of the client. This proximal term uses the $\ell_2$ norm to explicitly constrain the local model to be close to the latest global model, limiting the local update of non-IID clients to be not too far apart. However, this explicit constraint could inhibit FedProx from quickly finding a better model in the early stage. Similarly, based on multi-task learning, FedCurv \citep{shoham2019overcoming} adds penalty items for changes in important parameters related to other clients during local training. In addition, SCAFFOLD \citep{karimireddy2020scaffold} introduces control variables to correct the gradient of each local update and make it the same as the global update direction. In each round, the control variables are updated as the estimations of the difference between the update of the server model and the local model. However, these methods only implicitly reduce the impact of the inconsistency of objective, which is the fundamental dilemma of statistical heterogeneity, rather than eliminating it. This is exactly our motivation for proposing FedShift.

    In detail, we formulate the problem and propose our method FedShift in Section \ref{method}, which also includes the convergence analysis of FedShift under different assumptions and its superiority compared with FedAvg. In Section \ref{experiments}, we report our experimental results, we compare the accuracy and communication efficiency of our algorithm with other algorithms, and study the influence of different settings on the algorithm, such as degree of heterogeneity, the local epoch number and clients number. Finally, Section \ref{conclusion} concludes our paper.

    \section{FedShift: Federated Learning with Classifier Shift}
    \label{method}
    \subsection{Problem Formulation}
    \label{formulation}
    In federated learning, the global objective is to solve the following optimization problem:
    \begin{equation}\label{obj}
        \min _{\boldsymbol{w}}\left[L(\boldsymbol{w}) \triangleq  \sum_{i =1 }^{N} \frac{|\mathcal{D}_i|}{|\mathcal{D}|} L_{i}(\boldsymbol{w})\right],
    \end{equation}
    where $L_{i}(\boldsymbol{w})=\mathbb{E}_{(\boldsymbol{x}, y) \sim \mathcal{D}_{i}}\left[\ell_{i}(f(\boldsymbol{w} ;\boldsymbol{x}), y)\right]$ is the empirical loss of the $\mathit{i}$-th client that owns the local dataset $\mathcal{D}_{i}$, and $\mathcal{D} \triangleq \bigcup_{i=1}^N \mathcal{D}_{i}$ is a virtual entire dataset that includes all client‘s local data. $f(\boldsymbol{w} ;\boldsymbol{x})$ is the output of the model $\boldsymbol{w}$ when the input $\boldsymbol{x}$ is given, and $\ell_{i}$ denotes the loss function of the $i$-th client. Here, FL expects to learn a global model $\boldsymbol{w}$ that can perform well on the entire dataset $\mathcal{D}$.

    However, due to the inability to communicate local data, each client usually learns a local model $\boldsymbol{w}_i$ on its local dataset by minimizing the experience loss $L_{i}(\boldsymbol{w}_i)$. Then the server aggregates multiple local models to obtain a global model $\bar{\boldsymbol{w}}$. In FedAvg algorithm \citep{mcmahan2017communication}, each client adopts stochastic gradient descent (SGD) to update the local model $\boldsymbol{w}_i^{(t,\tau)}$  starting from $\boldsymbol{w}_{i}^{(t,0)} \triangleq \bar{\boldsymbol{w}}^{t-1}$ which is the latest global model. The local update process can be formulated as follows:
    \begin{equation}
        \boldsymbol{w}_{i}^{(t,\tau)}=\boldsymbol{w}_{i}^{(t,\tau-1)}-\eta \nabla_{\boldsymbol{w}} \ell_i(\boldsymbol{w}_{i}^{(t,\tau-1)},\mathcal{B}_{i}^{(t,\tau)})
    \end{equation}
    where $\eta$ is the client learning rate and $\boldsymbol{w}_{i}^{(t,\tau)}$ denotes the local model of client $i$ after the $\tau$-th local update in the $t$-th communication round. Also, $\ell_i(\boldsymbol{w}_{i}^{(t,\tau-1)},\mathcal{B}_{i}^{(t,\tau)}) \triangleq \sum_{(\boldsymbol{x}, y) \sim \mathcal{B}_{i}^{(t,\tau)}} \frac{1}{|\mathcal{B}_{i}^{(t,\tau}|} \left[\ell_{i}(f( \boldsymbol{w}_{i}^{(t,\tau-1)} ;\boldsymbol{x}), y)\right]$ where $\mathcal{B}_{i}^{(t,\tau)}$ represents the $\tau$-th mini-batch samples of the local dataset $\mathcal{D}_{i}$ in the $t$-th communication round.

    And then the server updates the global model by averaging the local model updates of all clients at the end of each communication round as:
    \begin{equation}
        \bar{\boldsymbol{w}}^{t} =  \bar{\boldsymbol{w}}^{t-1} + \sum_{i =1 }^{N} \frac{|\mathcal{D}_i|}{|\mathcal{D}|} (\boldsymbol{w}_{i}^{(t,\tau_i)}-\boldsymbol{w}_{i}^{(t,0)})
    \end{equation}
    where $\tau_i$ is the local iterations completed by client $i$ in the SGD optimizer with a fixed batch-size.

    \paragraph{Client drift} As mentioned in \citep{karimireddy2020scaffold,zhao2018federated}, the problem of client drift will occur during the federated learning process due to the statistical heterogeneity ($P_i(\boldsymbol{x},y) \neq P(\boldsymbol{x},y)$) where $P_i$ denotes the probability distribution for $(\boldsymbol{x},y)$ in local client and $P$ denotes the probability distribution for global data.

    Let $\boldsymbol{w}^{*}$ be the global optimum of $L(\boldsymbol{w})$ and $\boldsymbol{w}_{i}^{*}$ be the optimum of each client’s empirical loss $L_i(\boldsymbol{w})$. Actually, we have $\boldsymbol{w}_{i}^{*} \neq \boldsymbol{w}^{*}$ and $\sum_{i =1 }^{N} \frac{|\mathcal{D}_i|}{|\mathcal{D}|} \boldsymbol{w}_{i}^{*} \neq \boldsymbol{w}^{*}$ due to the heterogeneous data distribution and our equation \ref{obj}. Therefore, the direction of each local update of clients will deviate from the global update. This deviation is accumulated in multiple iterations of SGD, which will eventually lead to a drift between $\boldsymbol{w}^{t}$ (the true global update) and $\bar{\boldsymbol{w}}^t$ (the average of the client update aggregated by the server).

    \paragraph{Class Imbalance} For classification tasks, the statistical heterogeneity of federated learning is usually caused by class imbalance. Suppose the label space $ Y = [1,2,\dots,K]$, and $P(y)$ is the probability distribution of each class. The distribution of training data can be expanded as $P(\boldsymbol{x},y) = P(\boldsymbol{x}|y)P(y)$ and $P_i(\boldsymbol{x},y) = P_i(\boldsymbol{x}|y)P_i(y)$, where $P(\boldsymbol{x}|y)$ is the conditional probability distribution of class $y$. The subscript $i$ represents the data distribution and probability of client $i$. In many real-world application scenarios of federated learning, data collected by different clients (such as IoT cameras) usually has approximately the same conditional probability distribution of each class, which implies $P_i(\boldsymbol{x}|y) \approx P(\boldsymbol{x}|y)$. Therefore, the statistical heterogeneity of federated learning often appears as class imbalance, that is, $P_i(y) \neq P(y)$.

    \subsection{Method}
    \label{shift}
    In order to alleviate the degradation of model performance due to class imbalance, we propose FedShift which is shown in Algorithm \ref{fedshift}. We first start from some intuitions of our proposed method.

    Note that $\boldsymbol{w}_i^{*}\triangleq \arg\min_{\boldsymbol{w}} \mathbb{E}_{(\boldsymbol{x},y)\sim P_i} \left[ \ell_i ( f(\boldsymbol{w};\boldsymbol{x}),y) \right]$ is not the optimum of the global optimization problem ($\min_{\boldsymbol{w}} \mathbb{E}_{(\boldsymbol{x},y)\sim P} \left[ \ell ( f(\boldsymbol{w};\boldsymbol{x}),y) \right]$), because of the statistical heterogeneity ($P\neq P_i$), although the global evaluation function is consistent with the local evaluation function ($\ell_i=\ell$). An intuitive idea can be inspired, that is, resampling or reweighting the client data to make $P = P_i$. However, since there are few or no samples in some categories, this method does not achieve good results in practice. Some empirical results in Section \ref{experiments} show that reweighting is not effective and even bring a severe drop in accuracy compared to FedAvg.

    \begin{algorithm}[htbp]
        \label{fedshift}
        \caption{FedShift}
        \LinesNumbered 
        \KwIn{number of communication rounds T, number of clients N, the fraction of clients C, number of local epochs E, batch size B, learning rate $\eta$, the global label distribution $P(y)$ }
        \KwOut{the global model $\boldsymbol{w}^{T}$}
        \vspace{0.5em}
        initialize $\boldsymbol{w}^{0}$ \\
        $m \gets \max( \lfloor C*N \rfloor, 1)$ \\
        \For{communicate round $t = 0,1,2,\dots,T-1$}{
            $M_t \gets $ randomly select a subset containing $m$ clients \\
            \ForEach{client $i \in M_t$}{
                $\boldsymbol{w}_{i}^{t} = \boldsymbol{w}^{t}$ \\
                $\boldsymbol{w}_i^{t+1} \gets $ \bf{LocalUpdate}($\boldsymbol{w}_{i}^{t})$\\
            }
            $\boldsymbol{w}^{t+1} =\boldsymbol{w}^{t} + \sum_{i \in M_t } \frac{|\mathcal{D}_i|}{|\mathcal{D}|} (\boldsymbol{w}_{i}^{t+1}-\boldsymbol{w}_{i}^{t})  $\\
        }
        \vspace{1em}
        \bf{LocalUpdate} ($\boldsymbol{w}_{i}^{t}$):\\
        \For{epoch $e = 1,2,\dots,E$}{
            \ForEach{batch $\mathcal{B}_{i}^{t} = (\boldsymbol{x},y) \in \mathcal{D}_{i}$}{
                $ \tilde{\ell_i}(\boldsymbol{w}_i^{t},\mathcal{B}_{i}^{t}) = \sum_{(\boldsymbol{x}, y)} \frac{1}{|\mathcal{B}_{i}^{(t,k}|} \left[ \ell_i\left( f(\boldsymbol{w}_i^{t};\boldsymbol{x}) + \boldsymbol{s}_i,y \right)\right]$ \tcp*{$s_i$ follows Eq.\ref{si}}
                $ \boldsymbol{w}_{i}^{t}=\boldsymbol{w}_{i}^{t}-\eta \nabla  \tilde{\ell_i}(\boldsymbol{w}_i^{t},\mathcal{B}_{i}^{t}) $\\
            }
        }
        return $\boldsymbol{w}_i^{t}$\\
    \end{algorithm}

    Different from reweighting, FedShift modifies the local optimization objective of each client to satisfy that $ \tilde{\boldsymbol{w}_i^*} = \arg\min_{\boldsymbol{w}} \mathbb{E}_{(\boldsymbol{x},y)\sim P_i} \left[ \tilde{\ell_i} ( \boldsymbol{f}(\boldsymbol{w};\boldsymbol{x}),y) \right]$ also is the optimum of the global optimization problem $\min_{\boldsymbol{w}} \mathbb{E}_{(\boldsymbol{x},y)\sim P} \left[ \ell ( \boldsymbol{f}(\boldsymbol{w};\boldsymbol{x}),y) \right]$. Let $\tilde{\ell_i}$ denote the modified local optimization objective of client $i$. In FedShift, we add the shift $s_i$ on the classifier output of the model to modify the local optimization objective of client $i$, shown as:
    \begin{equation}
        \label{newf}
        \tilde{\ell_i} = \ell_i( \tilde{\boldsymbol{f}}(\boldsymbol{w}_i^{t};\boldsymbol{x}),y) = \ell_i\left( \boldsymbol{f}(\boldsymbol{w}_i^{t};\boldsymbol{x}) + \boldsymbol{s}_i,y \right)
    \end{equation}
    The shift $\boldsymbol{s}_i = [s_{i,1}, s_{i,2}, \dots, s_{i,K}]$ is calculated by the local category probability to the classifier at the end of network, as follows:
    \begin{equation}
        \label{si}
        \boldsymbol{s}_{i,k} = \ln(\frac{P_i(y=k)}{P(y=k)}) \qquad k=1,2,\dots,K
    \end{equation}
    where $P(y=k)=\sum_{i=1}^{N}\frac{|\mathcal{D}_i|}{|\mathcal{D}|}P_i(y=k)$. \footnote{The probability can be calculated using the secure aggregation algorithm \citep{bonawitz2016practical} without leaking any client information at the beginning of the entire learning process. More specifically, we use Laplace smoothing on each client to approximate the probability by frequency to guarantee secure computation of the class probability.} Then, we propose our following Theorem 1 to show the advantages of our FedShift theoretically.

    \begin{thm}
        \label{thm1}
        \textit{For FedShift, by add shift $s_i$ in the output of model, the local optimum $\boldsymbol{w}_i^*$ satisfies the global optimum of $\min_{\boldsymbol{w}} \mathbb{E}_{(\boldsymbol{x},y)\sim P} \left[ \ell ( f(\boldsymbol{w};\boldsymbol{x}),y) \right]$.}
    \end{thm}

    \textbf{Proof:} For classification, cross entropy loss is the most commonly used loss function, and the output of the neural network model usually passes through a softmax function to get the predicted category probability. Therefore, by definition, we have
    \begin{equation}
        \boldsymbol{w}_{i}^{*} = \arg\min_{\boldsymbol{w}} \mathbb{E}_{(\boldsymbol{x},y)\sim P_i} \left[ \ell_i ( \tilde{f}(\boldsymbol{w};\boldsymbol{x}),y) \right] = \arg\min _{\boldsymbol{w}} \mathbb{E}_{(\boldsymbol{x},y) \sim P_i}\left[-\sum_{k=1}^{K} \mathbbm{1}_{y=k} \log \frac{e^{\tilde{f}_k(\boldsymbol{w};\boldsymbol{x})}} {\sum_{j=1}^{K}e^{\tilde{f}_j(\boldsymbol{w};\boldsymbol{x})}}  \right]
    \end{equation}
    Let $q_i(y=k|\boldsymbol{x};\boldsymbol{w}) \triangleq \frac{e^{\tilde{f}_k(\boldsymbol{w};\boldsymbol{x})}} {\sum_{j=1}^{K}e^{\tilde{f}_j(\boldsymbol{w};\boldsymbol{x})}}$, we can derive that the optimal model $\boldsymbol{w}_i^{*}$ should satisfy $q_i(y=k|\boldsymbol{x};\boldsymbol{w}_{i}^{*}) = P_i(y=k|\boldsymbol{x})$. And according to Bayes' theorem, we have $ p_i(y=k|\boldsymbol{x})  = \frac{p_i(\boldsymbol{x}|y=k)p_i(y=k)}{\sum_{j=1}^{K}p_i(\boldsymbol{x}|y=j)p_i(y=j)} $, then we have
    \begin{equation}
        \tilde{f}_k(\boldsymbol{w}_{i}^{*};\boldsymbol{x}) = \ln(p_i(\boldsymbol{x}|y=k)p_i(y=k))+const , k=1,2,\dots,K
    \end{equation}
    Then, we consider the origin output which is added the classify shift $s_i$ in client $i$ by equation \ref{newf}, we have
    \begin{equation}
        \begin{aligned}
            f_k(\boldsymbol{w}_i^{*};\boldsymbol{x}) & = \tilde{f}_k(\boldsymbol{w}_i^{*};\boldsymbol{x}) - s_{i.k} \\
            {} & =\ln(p_i(\boldsymbol{x}|y=k)p_i(y=k))-\ln(\frac{p_i(y=k)}{p(y=k)})+const  \\
            {} & =\ln(p(\boldsymbol{x}|y=k)p(y=k))+const  \\
        \end{aligned}
    \end{equation}
    \begin{equation}
        q(y=k|\boldsymbol{x};\boldsymbol{w}_i^{*}) = \frac{e^{f_k(\boldsymbol{w}_i^{*};\boldsymbol{x})}} {\sum_{j=1}^{K}e^{f_j(\boldsymbol{w}_i^{*};\boldsymbol{x})}} = p(y=k|\boldsymbol{x}), k=1,2,\dots,K
    \end{equation}
    which means that $\boldsymbol{w}_{i}^{*}$ satisfies $\min_{\boldsymbol{w}} \mathbb{E}_{(\boldsymbol{x},y)\sim P} \left[ \ell ( f(\boldsymbol{w};\boldsymbol{x}),y) \right]$, which is the global optimum. $\hfill\qedsymbol$

    Note that FedProx can also be considered to have made such a modification,that is $\tilde{\ell_i}(\tilde{\boldsymbol{f}}(\boldsymbol{w};\boldsymbol{x}),y) = \ell_i(\boldsymbol{f}(\boldsymbol{w};\boldsymbol{x}),y)+\lambda\|\boldsymbol{w}-\bar{\boldsymbol{w}}\|_2^2$. but it does not guarantee that this modification has the properties shown in Theorem \ref{thm1}.

    FedShift is designed as a simple and effective approach based on FedAvg, only introducing lightweight but novel modifications in the local training phase. Benefiting from the lightweight modifications in local training, FedShift will not damage the data privacy and add any communication cost, which potentially can be combined with other aggregation optimization approaches.


    \subsection{Convergence Analysis}
    \label{proof}
    Properties outlined in Theorem \ref{thm1} motivate our FedShift convergence analysis. We will present theoretical results for strongly convex and non-convex functions. We first give some common assumptions about the function ${L}_i$ and $\nabla  \ell_i(\boldsymbol{w},\mathcal{B}_{i}) $, which is the unbiased stochastic gradient of ${L}_i$.
    \begin{assumption}
        \label{assumption1}
        \textit{For all $i$, ${L}_i$ has the properties of $\mu$-strong convexity and $\beta$-smooth:}
        \begin{equation*}
            \mu\text{-strongly convex: } {L}_i(\boldsymbol{v}) \geq {L}_i(\boldsymbol{w})+\langle(\boldsymbol{v}-\boldsymbol{w}), \nabla {L}_i(\boldsymbol{w})\rangle + \frac{\mu}{2}||\boldsymbol{v}-\boldsymbol{w}||_2^2
        \end{equation*}
        \begin{equation*}
            \beta\text{-smooth: } {L}_i(\boldsymbol{v}) \leq {L}_i(\boldsymbol{w})+\langle(\boldsymbol{v}-\boldsymbol{w}), \nabla {L}_i(\boldsymbol{w})\rangle + \frac{\beta}{2}||\boldsymbol{v}-\boldsymbol{w}||_2^2
        \end{equation*}
    \end{assumption}
    \begin{assumption}
        \label{assumption2}
        \textit{Bounded variances and second moments: There exits constants $\sigma > 0$ and $G > 0$ such that}
        \begin{equation*}
            \mathbb{E}_{\mathcal{B}_{i} \sim \mathcal{D}_{i}}\left\|\nabla \ell_{i}\left(\boldsymbol{w} ; \mathcal{B}_{i}\right)-\nabla {L}_{i}(\boldsymbol{w})\right\|_2^{2} \leq \sigma^{2}, \forall \boldsymbol{w}, \forall i
        \end{equation*}
        \begin{equation*}
            \mathbb{E}_{\mathcal{B}_{i} \sim \mathcal{D}_i} \left[ \| \nabla  \ell_i(\boldsymbol{w},\mathcal{B}_{i}) \|_2^2\right] \leq G^2, \forall \boldsymbol{w}, \forall i
        \end{equation*}
    \end{assumption}
    Then, we give a lemma about the gap between the local model and the local optimum as follows, where the detailed proof is in Appendix A.
    \begin{lem}
        \label{lem1}
        \textit{Under Assumption \ref{assumption1} and \ref{assumption2}, we have} $ \mathbb{E}(||\boldsymbol{w}_{i}^{t+1}-\boldsymbol{w}_{i}^{*}||_2^2) \leq (1-\eta\mu)^{I+1} ||\bar{\boldsymbol{w}}^{t}-\boldsymbol{w}_{i}^{*}||_2^2 + \frac{\eta}{\mu}G^2$, \textit{where $I$ denotes the iterations of SGD for each client in each rounds.}
    \end{lem}

    \begin{thm}
        \label{FedShiftthm}
        \textit{Under Assumption \ref{assumption1} and \ref{assumption2}, in FedShift, we have $ \mathbb{E}(||\bar{\boldsymbol{w}}^{t+1}-\boldsymbol{w}^{*}||_2^2) \leq (1-\eta\mu)^{(I+1)t} ||\bar{\boldsymbol{w}}^{0}-\boldsymbol{w}^{*}||_2^2 + \frac{\eta\left[1-(1-\eta\mu)^{(I+1)(t+1)}\right]}{\mu\left[1-(1-\eta\mu)^{I+1}\right]}G^2  $, where $\bar{\boldsymbol{w}}^{t+1}\triangleq \sum_{i =1}^{N}\frac{\boldsymbol{w}_{i}^{t+1}}{N}$ and $\bar{\boldsymbol{w}}^{0}$ is the initial global model.}
    \end{thm}

    \textbf{Proof:} Following Theorem \ref{thm1} and the strongly convex of $\mathcal{L}$, we can derive that $\boldsymbol{w}_i^*=\boldsymbol{w}^*$. Then, we have
    \begin{equation}
        \begin{aligned}
            \mathbb{E}(\|\bar{\boldsymbol{w}}^{t+1}-\boldsymbol{w}^{*}\|_2^2)   & = \mathbb{E}(\|\sum_{i =1}^{N}\frac{\boldsymbol{w}_{i}^{t+1}}{N}-\boldsymbol{w}^{*}\|_2^2) = \mathbb{E}( \| \frac{1}{N}\sum_{i=1}^{N}(\boldsymbol{w}_{i}^{t+1}-\boldsymbol{w}^{*}) \|_2^2) &&\\
            & \stackrel{(a)}{\leq} \frac{1}{N}\sum_{i=1}^{N} \mathbb{E}( \| (\boldsymbol{w}_{i}^{t+1}-\boldsymbol{w}_{i}^{*}) \|_2^2)
            \stackrel{(b)}{\leq} (1-\eta\mu)^{I+1} \|\bar{\boldsymbol{w}}^{t}-\boldsymbol{w}_{i}^{*}\|_2^2 + \frac{\eta}{\mu}G^2  &&\\
            & \stackrel{(recurrence)}{\leq} (1-\eta\mu)^{(I+1)t} ||\bar{\boldsymbol{w}}^{0}-\boldsymbol{w}^{*}||_2^2 + \frac{\eta\left[1-(1-\eta\mu)^{(I+1)(t+1)}\right]}{\mu\left[1-(1-\eta\mu)^{I+1}\right]}G^2 &&\\
        \end{aligned}
    \end{equation}
    where (a) follows from the Jensen's Inequality and $\boldsymbol{w}_i^*=\boldsymbol{w}^*$, (b) follows from Lemma \ref{lem1}.$\hfill\qedsymbol$

    Theorem \ref{FedShiftthm} shows us that under the strongly convex assumption of the function, benefiting from the classifier shift in FedShift, the global model can converge to the global optimum when there are enough iterations and communication rounds and a decayed learning rate.

    However, since Theorem \ref{thm1} does not hold on FedAvg, FedAvg does not have such good properties. We can get a lower bound of the gap between the global model and the global optimum in FedAvg, expressed as Theorem \ref{avgthm}.

    \begin{thm}
        \label{avgthm}
        \textit{For FedAvg, in the case of non-IID client data, there is a gap between the local optimal and the global optimal. Mark $\bar{\boldsymbol{w}}^{*}\triangleq \sum_{i=1}^{N}\frac{\boldsymbol{w}_i^{*}}{N}$. If we assume that $||\bar{\boldsymbol{w}}^{*}-\boldsymbol{w}^{*}||_2=\delta>0$, $||\boldsymbol{w}_i^{*}-\boldsymbol{w}^{*}||_2=\zeta>0$ and $||\bar{\boldsymbol{w}}^{0}-\boldsymbol{w}^{*}||_2=\gamma >0$, then under Assumption \ref{assumption1} and \ref{assumption2}, we have $ \mathbb{E}(||\bar{\boldsymbol{w}}^{t+1}-\boldsymbol{w}^{*}||_2^2) \geq \frac{\delta^2}{2}$ when $I$ satisfies that $I \geq \max\left\{ \frac{\ln(\frac{\frac{\delta^2}{16}-\frac{\eta}{\mu}G^2}{(\frac{5\delta}{4} +\zeta)^2})}{\ln(1-\eta\mu)} -1, \frac{\ln(\frac{\frac{\delta^2}{16}-\frac{\eta}{\mu}G^2}{(\zeta+\gamma)^2})}{\ln(1-\eta\mu)} -1 \right\}$.}
    \end{thm}
    \textbf{Proof:} Considering the first local update, from Lemma \ref{lem1}, we can get that $\mathbb{E}(\| \boldsymbol{w}_{i}^{1} - \boldsymbol{w}_{i}^{*}\|_2^2) \leq (1-\eta\mu)^{I+1} \|\bar{\boldsymbol{w}}^{0}-\boldsymbol{w}_{i}^{*}\|_2^2 + \frac{\eta}{\mu}G^2 \leq (1-\eta\mu)^{I+1} \left( \|\bar{\boldsymbol{w}}^{0}-\boldsymbol{w}^{*}\|_2 + \|\boldsymbol{w}^{*}-\boldsymbol{w}_{i}^{*}\|_2 \right)^2+ \frac{\eta}{\mu}G^2 \leq (1-\eta\mu)^{I+1} \left( \gamma + \zeta \right)^2+ \frac{\eta}{\mu}G^2 \leq \frac{\delta^2}{16}$. Then, we do a mathematical induction proof for $\mathbb{E}(\| \boldsymbol{w}_{i}^{t} - \boldsymbol{w}_{i}^{*}\|_2^2)\leq \frac{\delta^2}{16}$, which holds at round t. Then, we can derive
    \begin{equation}
        \label{lemme2}
        \begin{aligned}
            \mathbb{E}(\| \boldsymbol{w}_{i}^{t+1} - \boldsymbol{w}_{i}^{*}\|_2^2)
            & \leq (1-\eta\mu)^{I+1} \|\bar{\boldsymbol{w}}^{t}-\boldsymbol{w}_{i}^{*}\|_2^2 + \frac{\eta}{\mu}G^2  \\
            & \leq (1-\eta\mu)^{I+1} \left( \|\bar{\boldsymbol{w}}^{t}-\boldsymbol{w}^{*}\|_2 + \|\boldsymbol{w}^{*}-\boldsymbol{w}_{i}^{*}\|_2  \right)^2+ \frac{\eta}{\mu}G^2 \\
            & \leq (1-\eta\mu)^{I+1} \left( \|\bar{\boldsymbol{w}}^{t}-\bar{\boldsymbol{w}}^{*}\|_2 + \|\bar{\boldsymbol{w}}^{*}-\boldsymbol{w}^{*}\|_2 +\zeta \right)^2+ \frac{\eta}{\mu}G^2 \\
            & \leq (1-\eta\mu)^{I+1} \left(  \frac{1}{N}\sum_{i =1}^N\|\boldsymbol{w}_{i}^{t}-\boldsymbol{w}_{i}^{*}\|_2 + \delta +\zeta \right)^2+ \frac{\eta}{\mu}G^2 \\
            & \leq (1-\eta\mu)^{I+1} \left(  \frac{\delta}{4} + \delta +\zeta \right)^2+ \frac{\eta}{\mu}G^2  \leq \frac{\delta^2}{16} \\
        \end{aligned}
    \end{equation}
    Then, we have
    \begin{equation}
        \begin{aligned}
            \mathbb{E}(\|\bar{\boldsymbol{w}}^{t+1}-\boldsymbol{w}^{*}\|_2^2)   & = \mathbb{E}(\|\bar{\boldsymbol{w}}^{t+1}-\bar{\boldsymbol{w}}^{*} + \bar{\boldsymbol{w}}^{*} - \boldsymbol{w}^{*} \|_2^2) \\
            & \stackrel{(c)}{\geq} \mathbb{E}( \| \bar{\boldsymbol{w}}^{t+1}-\bar{\boldsymbol{w}}^{*}\|_2 - \| \bar{\boldsymbol{w}}^{*}-\boldsymbol{w}^{*}\|_2 )^2 \\
            & = \mathbb{E}( \| \bar{\boldsymbol{w}}^{t+1}-\bar{\boldsymbol{w}}^{*}\|_2^2)  + \mathbb{E}( \| \bar{\boldsymbol{w}}^{*}-\boldsymbol{w}^{*}\|_2^2) - 2\delta \mathbb{E}(\| \bar{\boldsymbol{w}}^{t+1}-\bar{\boldsymbol{w}}^{*} \|_2)
            \\
            & \geq 0 + \delta^2-2\delta {\mathbb{E}(\| \bar{\boldsymbol{w}}^{t+1}-\bar{\boldsymbol{w}}^{*} \|_2)} \\
            & \stackrel{(d)}{\geq}  \delta^2-\frac{2\delta}{N}\sum_{i =1}^{N} \mathbb{E}(\| \boldsymbol{w}_{i}^{t+1}-\boldsymbol{w}_{i}^{*} \|_2) \\
            & \stackrel{(e)}{\geq}  \delta^2-2\delta\sqrt{\frac{\delta^2}{16}} = \frac{\delta^2}{2} >0
            \\
        \end{aligned}
    \end{equation}
    where (c,d,e) follow the Triangle Inequality, the Jensen's Inequality and equation \ref{lemme2} respectively.$\hfill\qedsymbol$

    Furthermore, we consider the convergence of FedShift in the non-convex case, expressed as Theorem \ref{nonvexthm}. The detailed proof is in Appendix A.

    \begin{thm}
        \label{nonvexthm}
        \textit{Under assumption \ref{assumption1} and \ref{assumption2}, and removing the $\mu$-strongly convex assumption, we have $\frac{1}{T} \sum_{t=1}^{T}\mathbb{E}(|| \nabla_{\boldsymbol{w}} L (\bar{\boldsymbol{w}}^{t-1}) ||_2^2) \leq \frac{2}{\eta T}({L} (\bar{\boldsymbol{w}}^{0})-{L} (\bar{\boldsymbol{w}}^{*}) ) + 4\eta^2I^2G^2\beta^2+\frac{\beta}{N}\eta \sigma^2$.}
    \end{thm}

    \section{Experiments}
    \label{experiments}
    \subsection{Experimental Setup}
    \label{experimental setup}
    \paragraph{Datasets and Models} We conduct experiments on three public datasets including Cifar10 (60,000 images with 10 classes) \citep*{krizhevsky2009learning}, Cinic10 (270,000 images with 10 classes) \citep{darlow2018cinic}, and Tiny-Imagenet (100,000 images with 200 classes) \citep{Le2015TinyIV}. We follow the setting in \citep{wang2019federated,yurochkin2019bayesian}
    to generate the non-IID data partition by using Dirichlet distribution. Specifically, for class $c$, we sample $\mathit{p}_c \sim Dir_{N}(\alpha)$, where $\mathit{p}_{c,i}$ represents the proportion of data with category $k$ allocated to client $i$. The smaller $\alpha$ means the heavier statistical heterogeneity. For all experiments unless there are special instructions, we set $\alpha = 0.1$ and the number of clients $N = 10$ by default. In order to show that the algorithm is feasible on the actual deep learning model, we use ResNet18 \citep{he2016deep} as our network architecture for Cifar10 and Cinic10. For Tiny-Imagenet, we use ResNet50 \citep{he2016deep} to deal with more complex data.
    \paragraph{Baselines}
    We compare FedShift with three state-of-the-art approaches which are the most relevant to us, including FedAvg \citep{mcmahan2017communication}, FedProx \citep{MLSYS2020_38af8613}, and SCAFFOLD \citep{karimireddy2020scaffold} on all three datasets.
    \paragraph{Implementation}
    We use PyTorch \citep{paszke2019pytorch} to implement FedShift and the other baselines. We use the SGD with momentum as our optimizer for all experiments, where the SGD weight decay is set to $0.0001$ and the momentum to $0.9$. We adjust batchsize $B = 40$ and the learning rate $\eta = 0.01$ with a decay rate $0.95$ for every $10$ communication rounds. We take the best performance of each method for comparison.

    \subsection{Accuracy Comparison}
    For each dataset, we tune the number of local epochs $E$ from $\{1, 5, 10, 20\}$ based on FedAvg, and choose the best $E$ as the hyperparameter of other algorithms. The best $E$ for Cifar10, Cinic10, and Tiny-Imagenet are $5$, $1$, and $1$, respectively. Besides, for FedProx, we tune $\lambda$ from $\{0.001, 0.01, 0.1\}$, which is a hyperparameter to control the weight of its proximal term. The best $\lambda$ of FedProx for Cifar10, Cinic10, and Tiny-Imagenet are $0.01$, $0.001$, and $0.01$, respectively. Unless explicitly specified, we use $E$ and $\lambda$ for all the remaining experiments. The number of communication rounds is set to $100$ for Cifar10, $150$ for Cinic10 and $50$ for Tiny-Imagenet, where all federated learning approaches have little or no accuracy gain with more communications.

    \begin{table}[htbp]
        \caption{The accuracy of Reweight, FedShift and three baselines (FedAvg, FedProx and SCAFFOLD) on three test datasets (Cifar10, Cinic10 and Tiny-Imagenet).}
        \label{acc-comparison}
        \centering
        \begin{tabular}{c|c|c|c}
            \toprule
            Methods     & Cifar10     & Cinic10 & Tiny-Imagenet  \\
            \midrule
            FedAvg & 78.94\% & 72.26\%  & 35.14\%   \\
            FedProx     & 79.33\% & 71.57\% & 36.16\%     \\
            SCAFFOLD     & 77.75\% & 73.22\% & 35.18\%  \\
            FedShift(ours) & \bf{83.52\%} & \bf{74.86\%} & \bf{36.61\%} \\
            Reweighting & 63.15\% & 30.06\% & 13.25\% \\
            \bottomrule
        \end{tabular}
    \end{table}

    Table \ref{acc-comparison} shows the test accuracy of all approaches with the above settings. Comparing different federated learning approaches, we can observe that FedShift is the best approach among all tasks, which even can outperform FedAvg by $4.58$\% accuracy on Cifar10. For FedProx and SCAFFOLD, they are only superior to FedAvg in specific datasets and do not have a significant improvement. Reweighting has much worse accuracy than other methods as we mentioned in Section \ref{method}.

    \subsection{Discussion of Communication Efficiency}
\begin{figure}[htbp]
        \centering
        \subfigure[Cifar10]{
            \label{level.sub.1}
            \includegraphics[width=0.32\linewidth]{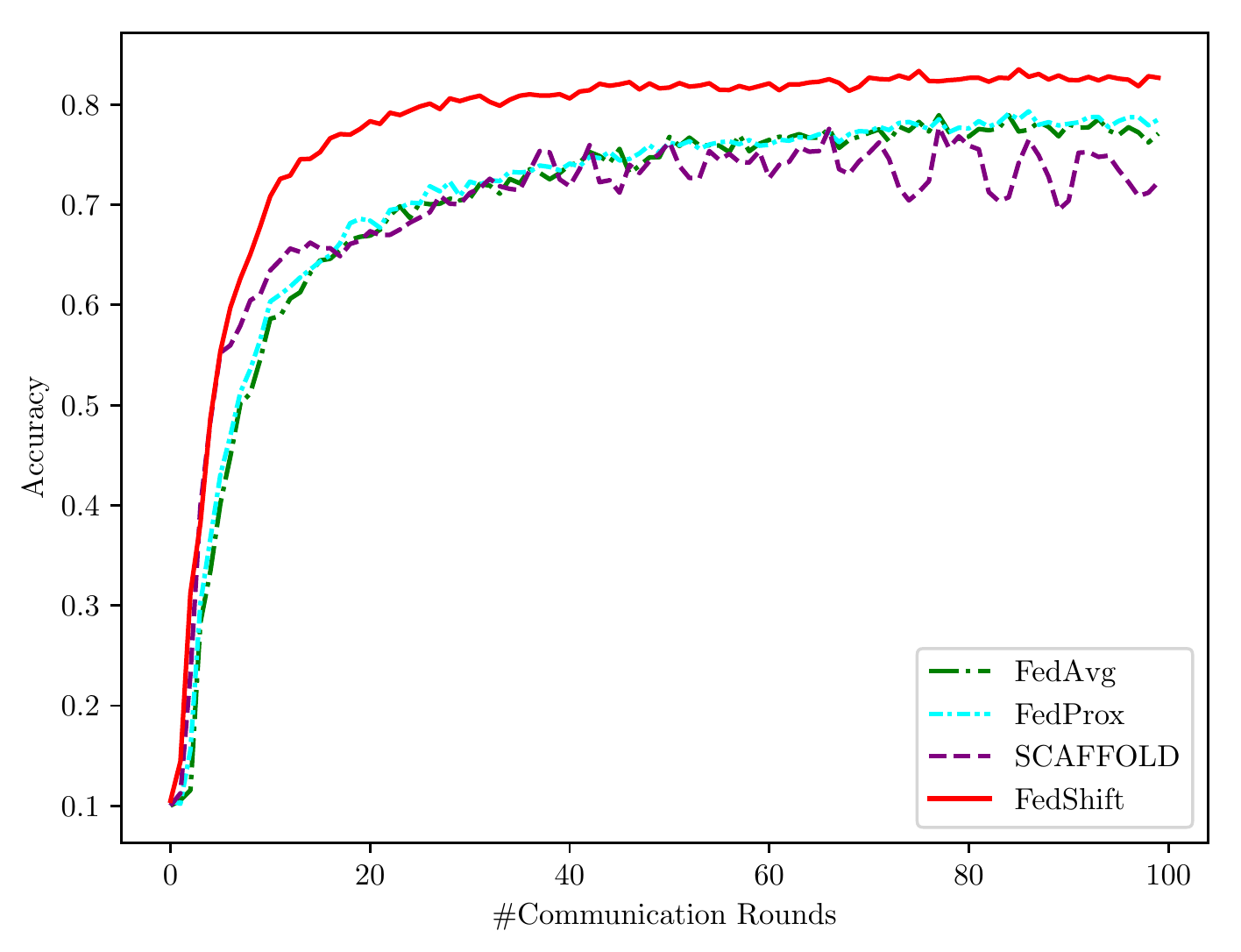}}
        \subfigure[Cinic10]{
            \label{level.sub.2}
            \includegraphics[width=0.32\linewidth]{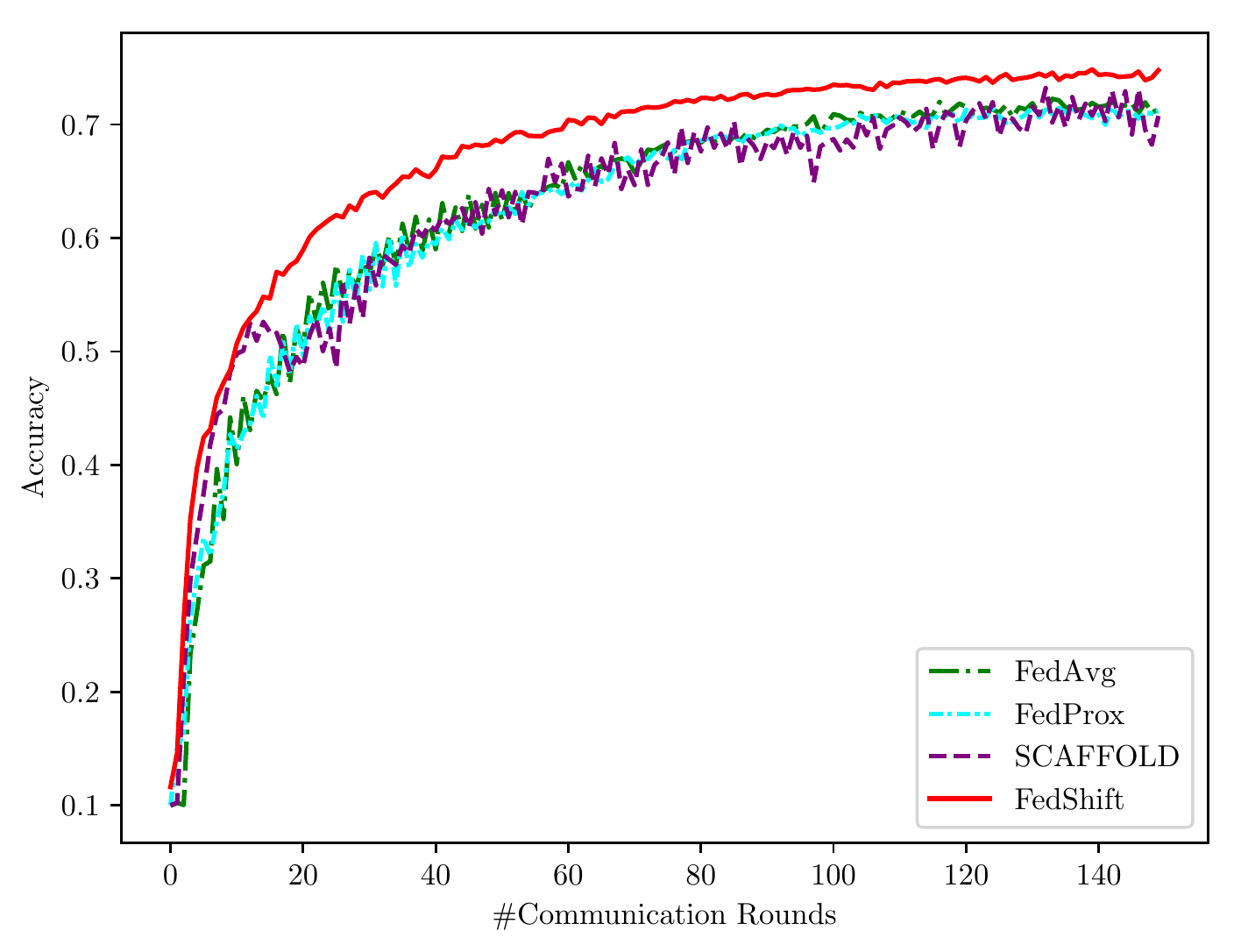}}
        \subfigure[Tiny-Imagenet]{
            \label{level.sub.3}
            \includegraphics[width=0.32\linewidth]{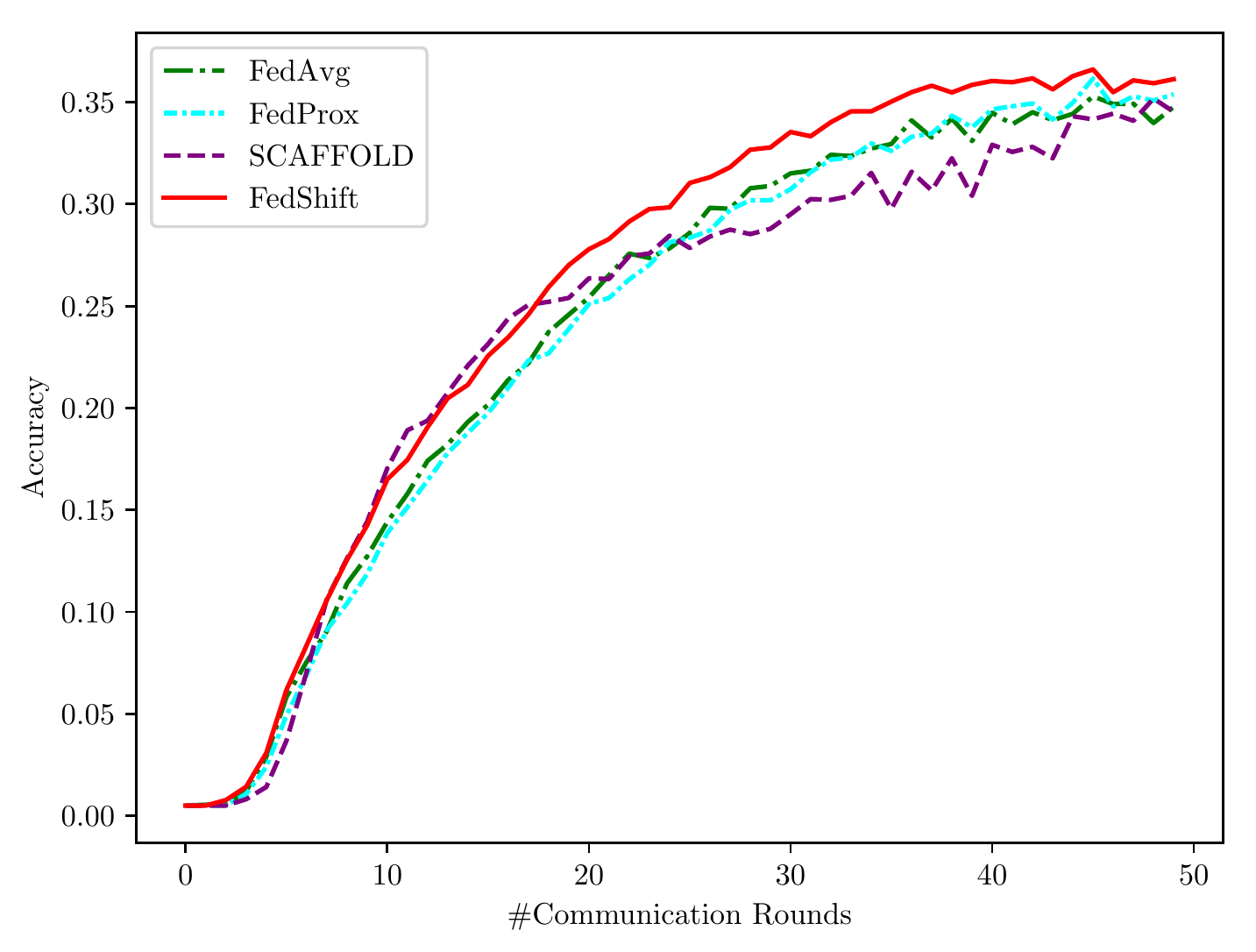}}
        \caption{The test accuracy in each communication round during training.}
        \label{accfig}
    \end{figure}

    Figure \ref{accfig} shows the accuracy in each communication round during training. As we can see, FedShift obviously has a faster convergence speed and higher accuracy compared with the other three methods. Moreover, unlike the other three methods, FedShift has a more stable upward curve due to the same optimization objective in all clients.

    In order to verify the communication efficiency of different approaches, we show the number of communication rounds to achieve the same accuracy for FedAvg in Table \ref{communication-rounds}. We can observe that the number of communication rounds is significantly reduced in FedShift. FedShift only needs less than a quarter of the number of rounds to reach the accuracy of FedAvg on the Cifar10. The speedup of FedShift is also significant on Cinic10 or Tiny-Imagenet. Therefore, we can consider FedShift is much more communication efficient than the other approaches.

    \begin{table}[htbp]
        \caption{The number of rounds of FedShift and three baselines (FedAvg,FedProx and SCAFFOLD) to achieve a consistent accuracy on three test datasets (Cifar10, Cinic10 and Tiny-Imagenet) respectively.}
        \label{communication-rounds}
        \centering
        \begin{tabular}{c|c|c|c|c|c|c}
            \toprule
            \multirow{2}{*}{Method} & \multicolumn{2}{|c|} { Cifar10 } & \multicolumn{2}{c|} { Cinic10 } & \multicolumn{2}{|c} { Tiny-Imagenet } \\
            {} & rounds & speedup & rounds & speedup & rounds & speedup \\
            \midrule
            FedAvg & 100 & $1 \times$ & 150 & $1 \times$ & 50 & $1 \times$ \\
            FedProx & $85$ & $1.18 \times$ & $\backslash$ & $<1 \times$ & 46 & $1.09 \times$ \\
            SCAFFOLD & $\backslash$ & $<1 \times$ & 133 & $1.13 \times$ & 49 & $1.02 \times$ \\
            FedShift (ours) & \bf{23} & $\mathbf{4.34} \times$ & \bf{81} & $\mathbf{1.85} \times$ & \bf{37} & $\mathbf{1.35} \times$ \\
            \bottomrule
        \end{tabular}
    \end{table}

    \subsection{Impact of Local Epochs}

    \begin{table}[htbp]
        \caption{The top-1 accuracy of FedShift and three baselines (FedAvg,FedProx and SCAFFOLD) on Cifar10 dataset with different number of local epochs.}
        \label{local-epochs}
        \centering
        \begin{tabular}{c|c|c|c|c}
            \toprule
            \cmidrule(r){1-2}
            Methods     & E=1     & E=5 & E=10 & E=20 \\
            \midrule
            FedAvg & 75.25\% & 78.94\%  & 76.47\% & 72.28\%  \\
            FedProx     & 75.31\% & 79.33\% & 77.27\% & 76.19\%    \\
            SCAFFOLD     & 70.60\% & 77.75\% & 77.95\% & 78.04\% \\
            FedShift (ours) & \bf{80.61\%} & \bf{83.52\%} & \bf{82.67\%} & \bf{81.63\%} \\
            \bottomrule
        \end{tabular}
    \end{table}

    We next focus on the effect of the number of local epochs on Cifar10. The results are shown in Table \ref{local-epochs}. When the number of local epochs is $1$, the local update is tiny, which leads to lower accuracy than more local epochs' results. However, FedShift still has the best accuracy. When the number of local epochs becomes too large, the accuracy of all approaches drops unless SCAFFOLD, which is due to the overfitting of local updates. Moreover, benefiting from the proximal term, FedProx performs better than FedAvg in all settings. Note that SCAFFOLD is far inferior to other algorithms when $E=1$, and has higher accuracy as the number of local epochs increases. Because the control variables in SCAFFOLD can be estimated more accurately with more local updates, SCAFFOLD has a higher tolerance for the number of local epochs. Nevertheless, FedShift clearly outperforms the other approaches. This further verifies that FedShift can effectively mitigate the negative effects of the accumulative client drift.

    \subsection{Impact of Data Heterogeneity}

    \begin{table}[htbp]
        \caption{The top-1 accuracy of FedShift and three baselines (FedAvg,FedProx and SCAFFOLD) on Cifar10 dataset with different parameter $\alpha$ of dirichlet distribution.}
        \label{alpha}
        \centering
        \begin{tabular}{c|c|c|c|c}
            \toprule
            \cmidrule(r){1-2}
            Methods  & $\alpha$=0.1 & $\alpha$=0.15 &$\alpha$=0.2 & $\alpha$=0.5 \\
            \midrule
            FedAvg  & 78.94\% & 80.11\% & 89.18\% & {91.27\%}  \\
            FedProx  & 75.95\% & 82.10\% & 89.43\% & 91.21\%    \\
            SCAFFOLD  & 77.75\% & 83.11\% & \bf{90.99}\% & \bf{92.22\%} \\
            FedShift (ours) & \bf{83.52\%} & \bf{86.21\%} & {90.76\%} & {91.26\%} \\
            \bottomrule
        \end{tabular}
    \end{table}

    Data heterogeneity is changed in this numerical study by varying the concentration parameter $\alpha$ of Dirichlet distribution on Cifar10. The results are shown in Table \ref{alpha}. For a smaller $\alpha$, the partition will be more unbalanced, we can significantly see the effectiveness of FedShift. When the unbalanced level decreases (i.e., $\alpha = 0.5$), all approaches have similar accuracy, and the control variable in SCAFFOLD actually degenerates into more considerable momentum to obtain higher accuracy.

    \subsection{Impact of the Number of Clients}

    \begin{table}[htbp]
        \caption{The top-1 accuracy of FedShift and three baselines (FedAvg,FedProx and SCAFFOLD) on Cifar10 dataset with different number of clients.}
        \label{clients}
        \centering
        \begin{tabular}{c|c|c|c}
            \toprule
            {Method} & { N=10,C=1.0 } & { N=20,C=0.5 } & { N=50,C=0.2 } \\
            \midrule
            FedAvg & 78.94\% & 79.49\% & 55.22\% \\
            FedProx & 79.33\% & 79.57\% & 56.24\%  \\
            SCAFFOLD & 77.75\% & 78.79\% & \bf{64.92\%} \\
            FedShift (ours) & \bf{83.52\%} & \bf{83.13\%} & 64.69\% \\
            \bottomrule
        \end{tabular}
    \end{table}

    To show the scalability of FedShift, we try more number of clients on Cifar10, including two settings: $20$ clients and $50$ clients. For better comparison, we adjust the proportion of clients participating in training in each round so that there are exactly $10$ clients each time, where $C=0.5/0.2$ for $N=20/50$. The communication round remains the same as the previous experiment, which is 100 rounds. The results are shown in Table \ref{clients}. FedShift achieves higher accuracy than FedAvg and FedProx in the different number of clients. Moreover, SCAFFOLD even outperforms FedShift with 50 clients and the fraction $C=0.2$ because of its estimation of the global gradient even if the gradients of some clients participating in training are small.
    \section{Conclusion}
    \label{conclusion}
    Focusing on the class imbalance in the statistical heterogeneity of federated learning, we propose FedShift in this paper, which is a simple and effective method that adds the shift on the classifier output based on the client class distribution in the local training phase. Then, we theoretically prove that the classifier shift in FedShift make the local optimal model satisfies the global optimum. Additionally, we prove the convergence of the FedShift algorithm and compare with FedAvg. We also conduct numerical studies, and the experimental results show that FedShift significantly outperforms the popular state-of-the-art algorithms on various datasets. Finally, as a future prospect, Fedshift has the potential to combine with the research of feature representation to deal with the inconsistency of the category conditional probabilities in each client, which is relaxation of our assumptions in our work.

%

    \bibliographystyle{plainnat}
    \bibliography{ref.bib}

\begin{thebibliography}{22}
\providecommand{\natexlab}[1]{#1}
\providecommand{\url}[1]{\texttt{#1}}
\expandafter\ifx\csname urlstyle\endcsname\relax
  \providecommand{\doi}[1]{doi: #1}\else
  \providecommand{\doi}{doi: \begingroup \urlstyle{rm}\Url}\fi

\bibitem[Bonawitz et~al.(2016)Bonawitz, Ivanov, Kreuter, Marcedone, McMahan,
  Patel, Ramage, Segal, and Seth]{bonawitz2016practical}
Keith Bonawitz, Vladimir Ivanov, Ben Kreuter, Antonio Marcedone, H~Brendan
  McMahan, Sarvar Patel, Daniel Ramage, Aaron Segal, and Karn Seth.
\newblock Practical secure aggregation for federated learning on user-held
  data.
\newblock \emph{arXiv preprint arXiv:1611.04482}, 2016.

\bibitem[Darlow et~al.(2018)Darlow, Crowley, Antoniou, and
  Storkey]{darlow2018cinic}
Luke~N Darlow, Elliot~J Crowley, Antreas Antoniou, and Amos~J Storkey.
\newblock Cinic-10 is not imagenet or cifar-10.
\newblock \emph{arXiv preprint arXiv:1810.03505}, 2018.

\bibitem[He et~al.(2016)He, Zhang, Ren, and Sun]{he2016deep}
Kaiming He, Xiangyu Zhang, Shaoqing Ren, and Jian Sun.
\newblock Deep residual learning for image recognition.
\newblock In \emph{Proceedings of the IEEE conference on computer vision and
  pattern recognition}, pages 770--778, 2016.

\bibitem[Ji et~al.(2019)Ji, Pan, Long, Li, Jiang, and Huang]{ji2019learning}
Shaoxiong Ji, Shirui Pan, Guodong Long, Xue Li, Jing Jiang, and Zi~Huang.
\newblock Learning private neural language modeling with attentive aggregation.
\newblock In \emph{2019 International Joint Conference on Neural Networks
  (IJCNN)}, pages 1--8. IEEE, 2019.

\bibitem[Ji et~al.(2021)Ji, Saravirta, Pan, Long, and Walid]{ji2021emerging}
Shaoxiong Ji, Teemu Saravirta, Shirui Pan, Guodong Long, and Anwar Walid.
\newblock Emerging trends in federated learning: From model fusion to federated
  x learning.
\newblock \emph{arXiv preprint arXiv:2102.12920}, 2021.

\bibitem[Kairouz et~al.(2019)Kairouz, McMahan, Avent, Bellet, Bennis, Bhagoji,
  Bonawitz, Charles, Cormode, Cummings, et~al.]{kairouz2019advances}
Peter Kairouz, H~Brendan McMahan, Brendan Avent, Aur{\'e}lien Bellet, Mehdi
  Bennis, Arjun~Nitin Bhagoji, Keith Bonawitz, Zachary Charles, Graham Cormode,
  Rachel Cummings, et~al.
\newblock Advances and open problems in federated learning.
\newblock \emph{arXiv preprint arXiv:1912.04977}, 2019.

\bibitem[Kaissis et~al.(2020)Kaissis, Makowski, R{\"u}ckert, and
  Braren]{kaissis2020secure}
Georgios~A Kaissis, Marcus~R Makowski, Daniel R{\"u}ckert, and Rickmer~F
  Braren.
\newblock Secure, privacy-preserving and federated machine learning in medical
  imaging.
\newblock \emph{Nature Machine Intelligence}, 2\penalty0 (6):\penalty0
  305--311, 2020.

\bibitem[Karimireddy et~al.(2020)Karimireddy, Kale, Mohri, Reddi, Stich, and
  Suresh]{karimireddy2020scaffold}
Sai~Praneeth Karimireddy, Satyen Kale, Mehryar Mohri, Sashank Reddi, Sebastian
  Stich, and Ananda~Theertha Suresh.
\newblock Scaffold: Stochastic controlled averaging for federated learning.
\newblock In \emph{International Conference on Machine Learning}, pages
  5132--5143. PMLR, 2020.

\bibitem[Krizhevsky et~al.(2009)Krizhevsky, Hinton,
  et~al.]{krizhevsky2009learning}
Alex Krizhevsky, Geoffrey Hinton, et~al.
\newblock Learning multiple layers of features from tiny images.
\newblock 2009.

\bibitem[Le and Yang(2015)]{Le2015TinyIV}
Ya~Le and X.~Yang.
\newblock Tiny imagenet visual recognition challenge.
\newblock 2015.

\bibitem[Li et~al.(2020)Li, Sahu, Zaheer, Sanjabi, Talwalkar, and
  Smith]{MLSYS2020_38af8613}
Tian Li, Anit~Kumar Sahu, Manzil Zaheer, Maziar Sanjabi, Ameet Talwalkar, and
  Virginia Smith.
\newblock Federated optimization in heterogeneous networks.
\newblock In I.~Dhillon, D.~Papailiopoulos, and V.~Sze, editors,
  \emph{Proceedings of Machine Learning and Systems}, volume~2, pages 429--450,
  2020.
\newblock URL
  \url{https://proceedings.mlsys.org/paper/2020/file/38af86134b65d0f10fe33d30dd76442e-Paper.pdf}.

\bibitem[Liu et~al.(2020)Liu, Huang, Luo, Huang, Liu, Chen, Feng, Chen, Yu, and
  Yang]{liu2020fedvision}
Yang Liu, Anbu Huang, Yun Luo, He~Huang, Youzhi Liu, Yuanyuan Chen, Lican Feng,
  Tianjian Chen, Han Yu, and Qiang Yang.
\newblock Fedvision: An online visual object detection platform powered by
  federated learning.
\newblock In \emph{Proceedings of the AAAI Conference on Artificial
  Intelligence}, volume~34, pages 13172--13179, 2020.

\bibitem[McMahan et~al.(2017)McMahan, Moore, Ramage, Hampson, and
  y~Arcas]{mcmahan2017communication}
Brendan McMahan, Eider Moore, Daniel Ramage, Seth Hampson, and Blaise~Aguera
  y~Arcas.
\newblock Communication-efficient learning of deep networks from decentralized
  data.
\newblock In \emph{Artificial Intelligence and Statistics}, pages 1273--1282.
  PMLR, 2017.

\bibitem[Paszke et~al.(2019)Paszke, Gross, Massa, Lerer, Bradbury, Chanan,
  Killeen, Lin, Gimelshein, Antiga, et~al.]{paszke2019pytorch}
Adam Paszke, Sam Gross, Francisco Massa, Adam Lerer, James Bradbury, Gregory
  Chanan, Trevor Killeen, Zeming Lin, Natalia Gimelshein, Luca Antiga, et~al.
\newblock Pytorch: An imperative style, high-performance deep learning library.
\newblock \emph{Advances in Neural Information Processing Systems},
  32:\penalty0 8026--8037, 2019.

\bibitem[Shoham et~al.(2019)Shoham, Avidor, Keren, Israel, Benditkis,
  Mor-Yosef, and Zeitak]{shoham2019overcoming}
Neta Shoham, Tomer Avidor, Aviv Keren, Nadav Israel, Daniel Benditkis, Liron
  Mor-Yosef, and Itai Zeitak.
\newblock Overcoming forgetting in federated learning on non-iid data.
\newblock \emph{arXiv preprint arXiv:1910.07796}, 2019.

\bibitem[Voigt and Von~dem Bussche(2017)]{voigt2017eu}
Paul Voigt and Axel Von~dem Bussche.
\newblock The eu general data protection regulation (gdpr).
\newblock \emph{A Practical Guide, 1st Ed., Cham: Springer International
  Publishing}, 10:\penalty0 3152676, 2017.

\bibitem[Wang et~al.(2019)Wang, Yurochkin, Sun, Papailiopoulos, and
  Khazaeni]{wang2019federated}
Hongyi Wang, Mikhail Yurochkin, Yuekai Sun, Dimitris Papailiopoulos, and
  Yasaman Khazaeni.
\newblock Federated learning with matched averaging.
\newblock In \emph{International Conference on Learning Representations}, 2019.

\bibitem[Wang et~al.(2020{\natexlab{a}})Wang, Yurochkin, Sun, Papailiopoulos,
  and Khazaeni]{wang2020federated}
Hongyi Wang, Mikhail Yurochkin, Yuekai Sun, Dimitris Papailiopoulos, and
  Yasaman Khazaeni.
\newblock Federated learning with matched averaging.
\newblock In \emph{International Conference on Learning Representations},
  2020{\natexlab{a}}.

\bibitem[Wang et~al.(2020{\natexlab{b}})Wang, Liu, Liang, Joshi, and
  Poor]{wang2020tackling}
Jianyu Wang, Qinghua Liu, Hao Liang, Gauri Joshi, and H~Vincent Poor.
\newblock Tackling the objective inconsistency problem in heterogeneous
  federated optimization.
\newblock \emph{Advances in Neural Information Processing Systems}, 33,
  2020{\natexlab{b}}.

\bibitem[Yeganeh et~al.(2020)Yeganeh, Farshad, Navab, and
  Albarqouni]{yeganeh2020inverse}
Yousef Yeganeh, Azade Farshad, Nassir Navab, and Shadi Albarqouni.
\newblock Inverse distance aggregation for federated learning with non-iid
  data.
\newblock In \emph{Domain Adaptation and Representation Transfer, and
  Distributed and Collaborative Learning}, pages 150--159. Springer, 2020.

\bibitem[Yurochkin et~al.(2019)Yurochkin, Agarwal, Ghosh, Greenewald, Hoang,
  and Khazaeni]{yurochkin2019bayesian}
Mikhail Yurochkin, Mayank Agarwal, Soumya Ghosh, Kristjan Greenewald, Nghia
  Hoang, and Yasaman Khazaeni.
\newblock Bayesian nonparametric federated learning of neural networks.
\newblock In \emph{International Conference on Machine Learning}, pages
  7252--7261. PMLR, 2019.

\bibitem[Zhao et~al.(2018)Zhao, Li, Lai, Suda, Civin, and
  Chandra]{zhao2018federated}
Yue Zhao, Meng Li, Liangzhen Lai, Naveen Suda, Damon Civin, and Vikas Chandra.
\newblock Federated learning with non-iid data.
\newblock \emph{arXiv preprint arXiv:1806.00582}, 2018.

\end{thebibliography}

\end{document}